\documentclass[conference]{IEEEtran}
\IEEEoverridecommandlockouts
\usepackage{cite}
\usepackage{amsmath,amssymb,amsfonts}
\usepackage{algorithmic}
\usepackage{graphicx}
\usepackage{textcomp}
\usepackage{xcolor}
\usepackage{booktabs}
\usepackage{CJKutf8}

\def\BibTeX{{\rm B\kern-.05em{\sc i\kern-.025em b}\kern-.08em
		T\kern-.1667em\lower.7ex\hbox{E}\kern-.125emX}}

\begin{document}
\begin{CJK*}{UTF8}{gbsn}
	
	\title{Semantic Similarity Matching for Patent Documents Using Ensemble BERT-related Model and Novel Text Processing Method\\ 
	}
	
        \author{
            \IEEEauthorblockN{Liqiang Yu}
            \IEEEauthorblockA{ \\
                \textit{The University of Chicago}\\
                Irvine, USA \\
                rexyu@outlook.com}
            \and
            \IEEEauthorblockN{Bo Liu\textsuperscript{*}}
            \IEEEauthorblockA{\textit{Software Engineering} \\
                \textit{Zhejiang University} \\
                Shanghai, China \\
                lubyliu45@gmail.com}
            \and
            \IEEEauthorblockN{Qunwei Lin}
            \IEEEauthorblockA{\textit{Information Studies} \\
                \textit{Trine University}\\
                Phoenix, USA \\
                linqunwei1030@outlook.com}
            \and
            \IEEEauthorblockN{Xinyu Zhao}
            \IEEEauthorblockA{\textit{Information Studies} \\
                \textit{Trine University}\\
                Phoenix, USA \\
                lution798@gmail.com}
            \and
            \IEEEauthorblockN{Chang Che}
            \IEEEauthorblockA{\textit{Mechanical engineering} \\
                \textit{The George Washington University}\\
                Atlanta, USA \\
                liamche1123@outlook.com}
        }

	\maketitle
	
	\begin{abstract}
        In the realm of patent document analysis, assessing semantic similarity between phrases presents a significant challenge, notably amplifying the inherent complexities of Cooperative Patent Classification (CPC) research. Firstly, this study addresses these challenges, recognizing early CPC work while acknowledging past struggles with language barriers and document intricacy. Secondly, it underscores the persisting difficulties of CPC research.
    
        To overcome these challenges and bolster the CPC system, This paper presents two key innovations. Firstly, it introduces an ensemble approach that incorporates four BERT-related models, enhancing semantic similarity accuracy through weighted averaging. Secondly, a novel text preprocessing method tailored for patent documents is introduced, featuring a distinctive input structure with token scoring that aids in capturing semantic relationships during CPC context training, utilizing BCELoss. Our experimental findings conclusively establish the effectiveness of both our Ensemble Model and novel text processing strategies when deployed on the U.S. Patent Phrase to Phrase Matching dataset.

	\end{abstract}
	
	
	\section{Introduction}
        In the realm of patent document analysis, the precise evaluation of semantic similarity between phrases poses a significant and fundamental challenge. This paper focuses on addressing this critical task, highlighting its particular relevance within the context of Cooperative Patent Classification (CPC). While early publications by Lent et al. \cite{lent1997discovering}, Larkey \cite{larkey1999patent}, and Gey et al. \cite{gey2001entry} laid the foundation for CPC, they also exposed limitations related to language barriers, precision, and adapting to the complexity of patent documents.

        Subsequent research efforts aimed to tackle these challenges by proposing innovative solutions. However, these solutions had their shortcomings. Chen and Chiu \cite{chen2013cross} focused on cross-language matching, with potential limitations in handling diverse patent document formats. Al-Shboul and Myaeng\cite{al2014wikipedia} employ Wikipedia for effective query expansion but face limitations with specialized technical terms. 
        
        Ever since the introduction of Deep Learning, Deep Learning-related techniques have seen extensive utilization in the field of CPC research. Prasad\cite{prasad2016searching} utilized CPC for bioremediation patent search, enhancing domain understanding. Shalaby et al.\cite{shalaby2018lstm} introduced LSTM, boosting patent classification accuracy and adaptability to changing taxonomies. Li et al.\cite{li2018deeppatent}, in their deep learning approach, demonstrated improved classification accuracy but required extensive computational resources.
        
        Moreover, the application of BERT-related techniques, as evidenced in the works by Lee and Hsiang \cite{lee2019patentbert}, and Bekamiri et al. \cite{bekamiri2021patentsberta}, significantly advanced CPC research by enhancing classification accuracy and efficiency. However, these methods still face challenges related to model scalability and data processing concerns.

        In the ever-evolving landscape of patent analysis and classification, the year 2023 has witnessed the emergence of significant research contributions. Yoo et al.\cite{yoo2023multi} delve into multi-label classification of Artificial Intelligence-related patents, employing advanced techniques. Ha and Lee\cite{ha2023examine} focus on evaluating the Cooperative Patent Classification (CPC) system, with a particular emphasis on patent embeddings. Hoshino et al.\cite{hoshino2023ipc} explore IPC prediction using neural networks and CPC's IPC classification. Additionally, Pais \cite{pais2023bert} investigates the CPC system's link to entity identification in patent text analysis. It is essential to acknowledge that these studies may exhibit certain limitations, offering opportunities for further research and refinement in the patent analysis field.
        
        To overcome these challenges and further enhance the capabilities of the CPC system, this paper introduces an ensemble approach. In contrast to the traditional methods mentioned earlier, the ensemble method leverages the strengths of multiple BERT-related models, including DeBERTaV3\cite{he2021debertav3} related models Microsoft's DeBERTa-v3-large, MoritzLaurer's DeBERTa-v3-large-mnli-fever-anli-ling-wanli, Anferico's BERT for patents\cite{devlin2018bert}, and Google's ELECTRA-large-discriminator\cite{clark2020electra}. This ensemble approach seeks to provide a comprehensive solution to the issues faced in previous research, thereby advancing the field of CPC.

        Our approach involves a novel text preprocessing method (V3) that groups and aggregates anchor and context pairs, resulting in each pair having an associated target list and score list. This structured input format adheres to a well-defined pattern, including tokens like [CLS], [SEP], and [TAR], designed to facilitate the model's understanding and analysis of patent document content. Our experiment results demonstrate the effectiveness of our Ensemble Model and novel text processing strategies when applied to the U.S. Patent Phrase to Phrase Matching dataset.
        The main contributions of this work can be summarized as follows:

        \begin{itemize}
            \item We proposed an ensemble of four deep learning models, including DeBERTaV3, Microsoft's DeBERTa-v3-large, MoritzLaurer's DeBERTa-v3-large-mnli-fever-anli-ling-wanli, Anferico's BERT for patents, and Google's ELECTRA-large-discriminator to enhances patent document analysis.
            \item We proposed a novel text preprocessing (V3) to group and aggregate anchor-context pairs, creating associated target and score lists.
            \item Our experiments confirm the efficacy of our Ensemble Model and novel text processing strategies on the U.S. Patent Matching dataset.
        \end{itemize}
        Moreover, this paper is structured as follows: The introduction sets the stage for understanding the challenges in CPC research. The related work section provides an overview of prior research efforts in the field. The algorithm and model section delves into the innovative ensemble approach and novel text preprocessing method. The conclusion section summarizes the contributions and the potential impact of this research on CPC analysis.

	\section{RELATED WORK}
        A number of initial publications established the groundwork for the Cooperative Patent Classification (CPC) system. Lent et al. \cite{lent1997discovering}  explored text data trends, relevant to CPC's patent document organization. Larkey \cite{larkey1999patent} contributed to patent search and classification, aligning with CPC's goal of effective categorization.

        CPC research has spanned language barriers, precision, and deep learning to advance patent classification and analysis. Notably, Al-Shboul and Myaeng's work \cite{al2014wikipedia} introduced "Wikipedia-based query phrase expansion" to enhance CPC's search precision and recall. 

        Due to the rapid advancements in deep learning, an increasing number of studies are being employed in the realm of CPC research. Prasad\cite{prasad2016searching} employed Cooperative Patent Classification (CPC) to conduct a comprehensive search for bioremediation patents, contributing to an enhanced understanding of the patent landscape in this domain. Shalaby et al.\cite{shalaby2018lstm} introduced an innovative method using Long Short-Term Memory (LSTM) networks enhanced the accuracy of patent classification, offering greater adaptability to changing patent taxonomies and more efficient patent organization and retrieval. Li et al.'s \cite{li2018deeppatent} "DeepPatent" with convolutional neural networks and word embeddings contributes to evolving and refining CPC's capabilities. Furthermore, studies enhanced CPC using BERT techniques, elevating patent document classification accuracy and efficiency. Lee and Hsiang fine-tuned a BERT model for patent classification in their pioneering work "PatentBERT" \cite{lee2019patentbert}.
        
        In the latest research conducted in 2023, the exploration of the Cooperative Patent Classification (CPC) system has continued to evolve. Yoo et al\cite{yoo2023multi}. examine multi-label classification of Artificial Intelligence-related patents, utilizing Modified D2SBERT and Sentence Attention mechanisms. Meanwhile, Ha and Lee's\cite{ha2023examine} article explores the effectiveness of the CPC system, focusing on patent embeddings. Hoshino et al.\cite{hoshino2023ipc} investigate IPC prediction using neural networks and CPC's IPC classification for patent document content. Additionally, Pais \cite{pais2023bert} delves into the CPC system's connection with entity linking in patent text analysis. Together, these studies significantly enhance our comprehension of the CPC system's role in patent analysis and classification, reflecting the latest advancements in the field.

	\section{ALGORITHM AND MODEL}
  
            \subsection{Ensemble Model} 
        To overcome the challenges inherent in patent document analysis and enhance the Cooperative Patent Classification (CPC) system's capabilities, we propose an innovative approach to improve patent document analysis and enhance the Cooperative Patent Classification (CPC) system. Instead of relying on single-model methods, our approach utilizes a diverse ensemble of deep learning models, including DeBERTaV3, Microsoft's DeBERTa-v3-large, MoritzLaurer's DeBERTa-v3-large-mnli-fever-anli-ling-wanli, Anferico's BERT for patents, and Google's ELECTRA-large-discriminator. Each model is chosen for its specific capabilities in capturing semantic relationships and nuances in patent documents.
        
        \begin{figure}[htbp]
        \centering
        \includegraphics[width=1\linewidth]{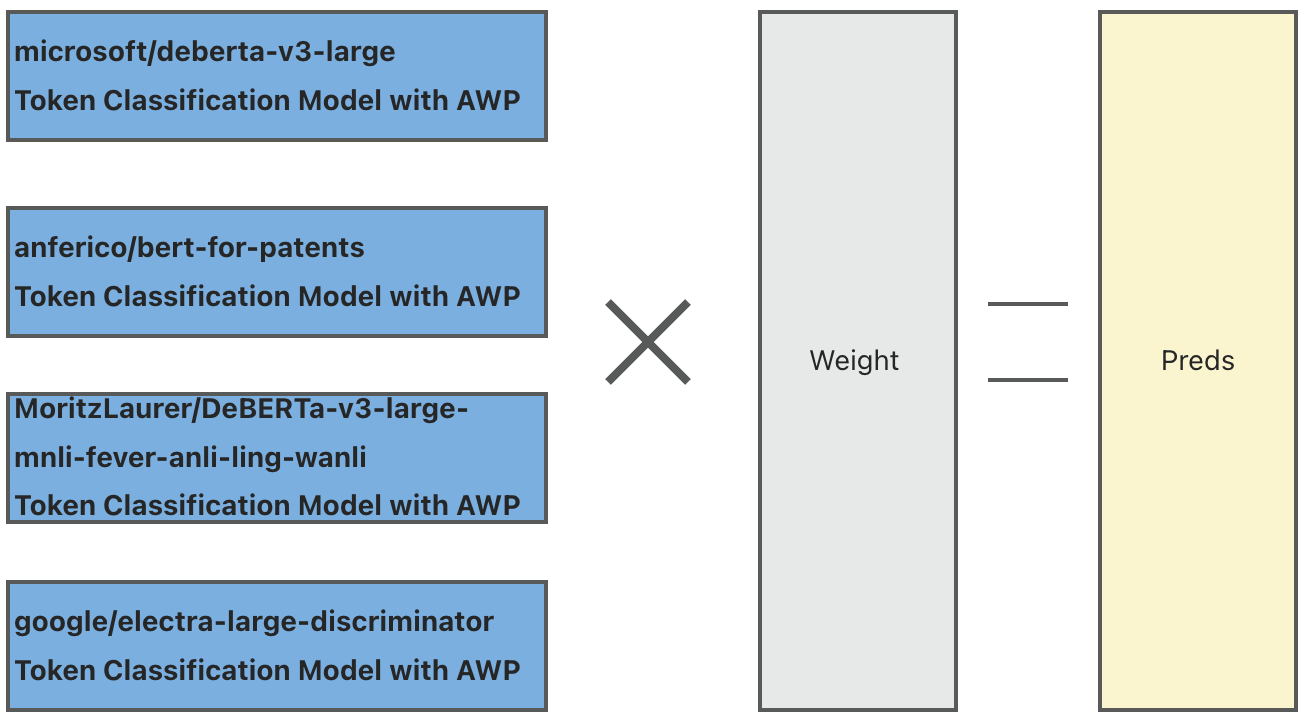}
        \caption{Ensemble Model}
        \label{fig:ensemble model}
        \end{figure}
        As shown in Fig.\ref{fig:ensemble model}, this ensemble model is designed to offer a comprehensive and fine-grained understanding of patent texts. The core principle of our ensemble model is the weighted averaging of predictions from individual models, where the weights are determined based on their performance on the validation data. Mathematically, this is represented as:

        \begin{equation}
        \hat{y}_e = \sum_{i=1}^{N} w_i \cdot \hat{y}_i 
        \end{equation}

        Where $\hat{y}_e$ represents the ensemble prediction, $\hat{y}_i$ represents the prediction from the $i$-th model, and $w_i$ represents the weight assigned to the $i$-th model. The weights $w_i$ are optimized through a validation process to maximize the ensemble's overall accuracy and semantic understanding of patent documents. This ensemble approach ensures that the CPC system benefits from the strengths of each individual model while mitigating their weaknesses, resulting in improved accuracy and efficiency in patent document analysis.

            \subsection{Novel Text Preprocessing Method}
\begin{figure}[htbp]
        \centering
        \includegraphics[width=1\linewidth]{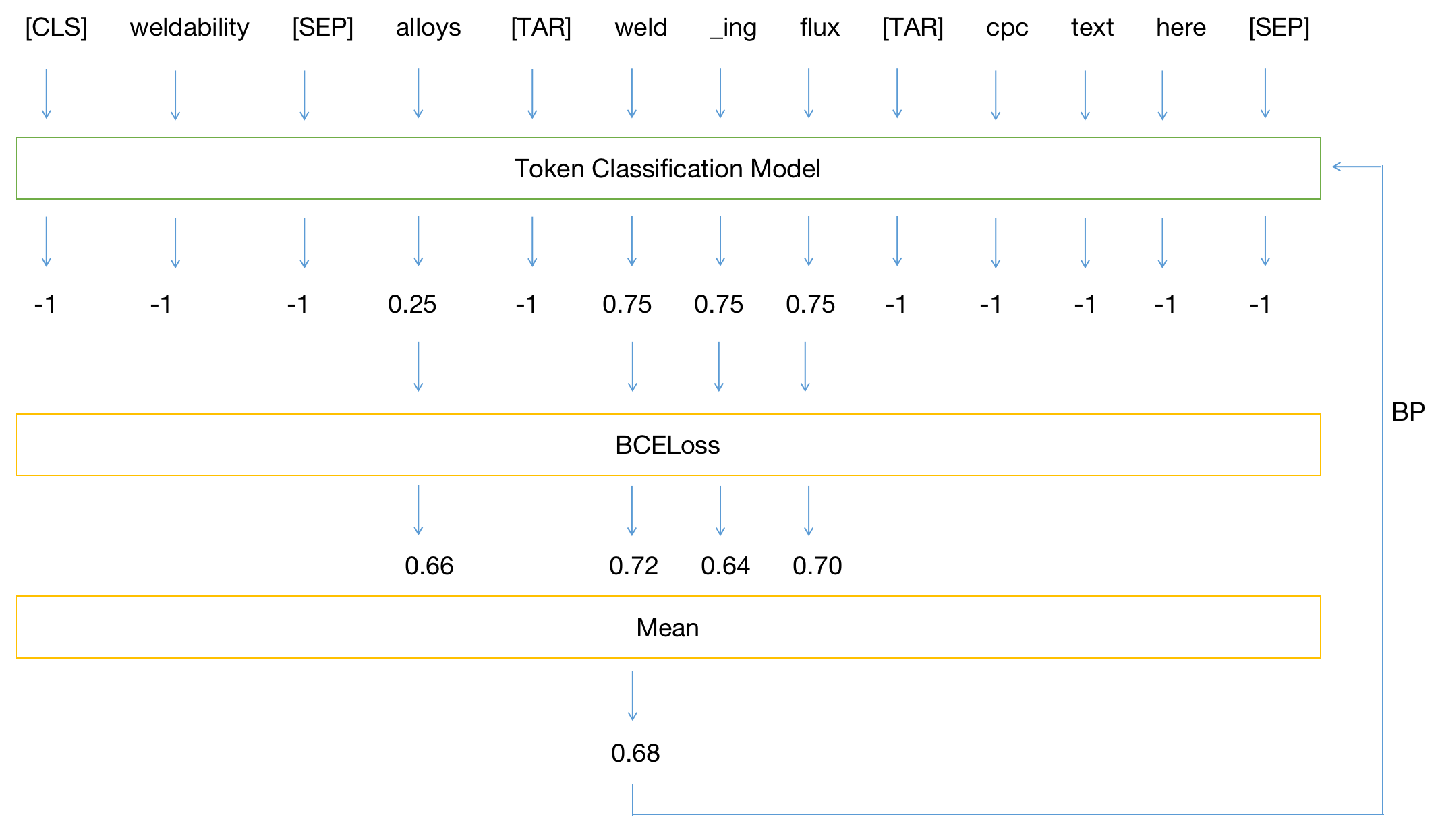}
        \caption{Data processing methods}
        \label{fig:text preprocessing}
        \end{figure}

        Our approach involves a meticulous text preprocessing method V3, where anchor and context pairs are thoughtfully grouped and aggregated, resulting in each pair having an associated target list and score list. This text preparation is essential for effectively assessing semantic similarity in patent documents. As shown in Fig.\ref{fig:text preprocessing}, the heart of our methodology lies in the structured input format we employ. This format adheres to a well-defined pattern, which includes tokens like [CLS], [SEP], and [TAR]. This structured input is designed to facilitate the model's understanding and analysis of patent document content.
    
        In our model, each token is assigned a score, a process efficiently executed within the TrainDataset class during data processing. This step ensures that the model can discern the significance of individual tokens within the Cooperative Patent Classification (CPC) context. The model's output is a sequence of the same length as the input, with each token receiving a predicted score. Even non-target tokens, such as [CLS], [SEP], and [TAR], receive scores, albeit with a true score set to -1, as they are not directly relevant to the semantic similarity assessment.
    
        For effective training and fine-tuning, we use BCELoss, a loss function comparing predicted scores to ground truth, aiming to align predicted and true scores, enhancing our model's patent document phrase similarity assessment. The loss function used is Binary Cross-Entropy Loss (BCELoss), which is defined as:
        \begin{equation}
        \mathcal{L} = -\frac{1}{N} \sum_{i=1}^{N} \left(G_i \cdot \log(P_i) + (1 - G_i) \cdot \log(1 - P_i)\right)
        \end{equation}
        Here, \(\mathcal{L}\) represents the overall loss for a batch of tokens, \(N\) is the total number of tokens in the batch, \(G_i\) is the ground truth score for token \(i\), and \(P_i\) is the predicted score for the same token. BCELoss guides the model during training to minimize the discrepancies between predicted and ground truth scores, facilitating the accurate assessment of semantic similarity between patent document phrases.

            \subsection{Datasets}
        The dataset provided for this task consists of pairs of phrases, which include an anchor phrase and a target phrase. The primary objective is to evaluate the degree of similarity between these phrases, utilizing a rating scale that ranges from 0 (indicating no similarity) to 1 (representing identical meaning). This assessment of similarity is unique in that it is conducted within the context of patent subject classification, specifically based on the Cooperative Patent 

        Scores in the dataset range from 0 to 1, with increments of 0.25, each representing a specific level of similarity. The entire dataset contains 48, 548 entries with 973 unique anchors, split into a training ($75\%$), validation ($5\%$), and test ($20\%$) sets. When splitting the data all of the entries with the same anchor are kept together in the same set. There are 106 different context CPC classes and all of them are represented in the training set.
                
            \subsection{Evalution metrics}

        The Pearson Correlation Coefficient ($r$) \cite{cohen2009pearson} is a statistical measure used to assess the strength and direction of the linear relationship between two variables. Its values lie within the range of -1 to 1.
        
        The evaluation metric used was the Pearson correlation coefficient ($r$) between the predicted and actual similarity scores, where a higher $r$ indicates a stronger linear relationship between predictions and ground truth scores. Submissions were assessed based on the Pearson correlation coefficient ($r$) between the predicted ($\hat{y}_i$) and actual ($y_i$) similarity scores, calculated as:
        
        \begin{equation}
        r = \frac{\sum \frac{(x_i - \overline{x})(y_i - \overline{y})}{n}}{\sqrt{\frac{\sum (x_i - \overline{x})^2}{n}} \sqrt{\frac{\sum (y_i - \overline{y})^2}{n}}}
        \end{equation}
        
        where $x_i$ and $y_i$ represent individual data points, $\overline{x}$ and $\overline{y}$ are the means of $x$ and $y$, respectively, and $n$ is the number of data set samples. The Pearson correlation coefficient measures the strength of the linear relationship between predicted and actual similarity scores, reflecting model performance in patent phrase similarity.

        We rigorously evaluated our model's performance and generalization with a 4-fold Cross-Validation \cite{browne2000cross} approach, maintaining label balance using MultiLabelStratifiedKFold. This method comprehensively assessed effectiveness across dataset subsets.
            
            \subsection{Results}
        In this section, we present the performance evaluation of our model variants (denoted as V1, V2, and V3) using the DeBERTa-v3-large architecture. We assessed the model's capabilities in U.S. Patent Phrase-to-Phrase Matching across different text processing strategies. Specifically, we considered the following variants:

        \begin{itemize}
            \item \textbf{V1}: The input text utilized the input structure: [CLS] anchor [SEP] target [SEP] context.
            \item \textbf{V2}:  The input text incorporated the input structure: [CLS] anchor [SEP] target [SEP] context [SEP] context...
            \item \textbf{V3}: As demonstrated in the section dedicated to Text Preprocessing Method.
        \end{itemize}
    
        The results are summarized in Table 1:

        \begin{table}[h]
            \centering
            \caption{Performance of Model Variants} 
            \label{tab:results}
            \begin{tabular}{cc}
                \toprule
                \textbf{Model Variant} & \textbf{CV Score} \\
                \midrule 
                V1 & 0.8347 \\
                V2 & 0.8369 \\
                V3 & 0.8512 \\
                \bottomrule 
            \end{tabular}
        \end{table}

        Our experiment results demonstrate the effectiveness of various text processing strategies, particularly highlighting the superior performance of text preprocessing method V3 among the tested approaches.

        Moreover, the ensemble model's hyperparameters were meticulously selected to optimize its performance in measuring semantic similarity within patent documents. These parameters included a maximum sequence length of 400, a learning rate of $1×10^{-5}$, attention weight perturbation epsilon of $1×10^{-2}$, an Adversarial Weight Perturbation (AWP)\cite{wu2020adversarial} learning rate of $1×10^{-4}$, a maximum gradient norm of 1000, epsilon of $1×10^{-5}$, and the utilization of Text Processing Strategy V3 with BCELoss and AWP. These settings were fine-tuned to ensure the ensemble model's effectiveness in combining the strengths of multiple deep learning models, resulting in an impressive ensemble score as previously discussed. The outcomes of the experiments are presented in Table 2, depicted as follows:
        \begin{table}[h]
            \centering
            \caption{Ensemble Model Results} 
            \label{tab:ensemble-results}
            \begin{tabular}{ccc}
                \toprule 
                \textbf{Model} & \textbf{Weight} & \textbf{CV Score} \\
                \midrule 
                Microsoft/DeBERTa-v3-large & 0.35 & 0.8512 \\
                Anferico/BERT-for-Patents & 0.2 & 0.8382 \\
                Google/ELECTRA-large & 0.25 & 0.8503 \\
                MoritzLaurer/DeBERTa-v3-large & 0.2 & 0.8385 \\
                \midrule 
                \multicolumn{2}{c}{Ensemble Model} & 0.8534 \\
                \bottomrule 
            \end{tabular}
        \end{table}
        
        The ensemble strategy incorporated these models with different weights to maximize their collective efficacy. The ensemble model's impressive performance was demonstrated by its Cross-Validation (CV) score, with Microsoft's DeBERTa-v3-large contributing a CV score of 0.8512, Anferico's BERT for patents with a CV score of 0.8382, Google's ELECTRA-large-discriminator scoring 0.8503 in CV, and MoritzLaurer's DeBERTa-v3-large-mnli-fever-anli-ling-wanli achieving a CV score of 0.8385. These models were blended with weights of 0.35, 0.2, 0.25, and 0.2, respectively, to create the ensemble. The final ensemble score, measured using the Pearson correlation coefficient, reached an impressive 0.8534, underscoring the success of this approach in enhancing semantic similarity measurement for patent documents. The table summarizes these results for clarity.

        \section{Conclusion}

        AI, notably in Bioinformatics, drives medical AI integration's rapid growth across diverse fields. Amidst the rapid advancement of artificial intelligence in diverse fields, our study delves into the intricate realm of semantic similarity assessment within patent documents, particularly in the context of the Cooperative Patent Classification (CPC) framework. While prior research laid the CPC foundation, it grappled with language barriers and precision issues. Subsequent innovative solutions faced constraints, and recent strides using BERT-related techniques showed promise but raised scalability and text processing concerns.
        
        To overcome these challenges and bolster the CPC system, our paper introduces an ensemble approach, harnessing multiple deep learning models, including DeBERTaV3-related ones, each meticulously trained with BCELoss. We also present creative data processing methods tailored to patent document nuances, featuring an innovative input structure that assigns scores to individual tokens. The incorporation of BCELoss during training leverages both predicted and ground truth scores, enabling fine-grained semantic analysis.
        
        By merging these innovations with traditional similarity assessment, our work aims to significantly enhance patent document analysis efficiency and precision. Our experimental findings conclusively establish the effectiveness of both our Ensemble Model and novel text processing strategies when deployed on the U.S. Patent Phrase to Phrase Matching dataset.

        \bibliographystyle{IEEEtran}
        \bibliography{references}
\end{CJK*}	
\end{document}